\documentclass[10pt,twocolumn,letterpaper]{article}

\usepackage{cvpr}              

\usepackage[dvipsnames]{xcolor}

\usepackage{amsmath}
\usepackage{amsfonts}
\usepackage{amssymb}
\usepackage{amsthm}
\usepackage{xcolor}
\usepackage{multicol}
\usepackage{multirow}
\usepackage{arydshln}
\usepackage{booktabs}
\usepackage{float}
\usepackage{graphicx}
\usepackage{bm}
\usepackage[accsupp]{axessibility}

\definecolor{cvprblue}{rgb}{0.21,0.49,0.74}
\usepackage[pagebackref,breaklinks,colorlinks,citecolor=cvprblue]{hyperref}

\def\taskname{generative identity unlearning}
\def\frameworkname{GUIDE}

\def\paperID{14335} 
\def\confName{CVPR}
\def\confYear{2024}

\title{Generative Unlearning for Any Identity}

\author{
{Juwon Seo}$^1$\thanks{Equal contribution}
\and
{Sung-Hoon Lee}$^1$\footnotemark[1]
\and
{Tae-Young Lee}$^1$\footnotemark[1]
\and
{Seungjun Moon}$^2$
\and
{Gyeong-Moon Park}$^1$\thanks{Corresponding author} \\
\\
$^1${Kyung Hee University, Yongin, Republic of Korea} \\
$^2${KLleon Tech., Seoul, Republic of Korea} 
\and
\tt{\small{\{jwseo001, sunghoonlee961, slcks1, gmpark\}@khu.ac.kr}}
\\
\tt{\small{seungjun.moon@klleon.io}}
}

\begin{document}
\maketitle
\begin{abstract}
\vspace{-0.1cm}
Recent advances in generative models trained on large-scale datasets have made it possible to synthesize high-quality samples across various domains.
Moreover, the emergence of strong inversion networks enables not only a reconstruction of real-world images but also the modification of attributes through various editing methods.
However, in certain domains related to privacy issues, e.g., human faces, advanced generative models along with strong inversion methods can lead to potential misuses.
In this paper, we propose an essential yet under-explored task called \taskname{}, which steers the model not to generate an image of specific identity.
In the \taskname{}, we target the following objectives:
(\lowercase{\romannumeral1}) preventing the generation of images with a certain identity, and (\lowercase{\romannumeral2}) preserving the overall quality of the generative model.
To satisfy these goals, we propose a novel framework, \textbf{G}enerative \textbf{U}nlearning for Any \textbf{IDE}ntity (\textbf{\frameworkname{}}), which prevents the reconstruction of a specific identity by unlearning the generator with only a single image.
\frameworkname{} consists of two parts:
(\lowercase{\romannumeral1}) finding a target point for optimization that un-identifies the source latent code and (\lowercase{\romannumeral2}) novel loss functions that facilitate the unlearning procedure while less affecting the learned distribution.
Our extensive experiments demonstrate that our proposed method achieves state-of-the-art performance in the generative machine unlearning task. 
The code is available at \href{https://github.com/KHU-AGI/GUIDE}{https://github.com/KHU-AGI/GUIDE}.
\end{abstract}    
\section{Introduction}
\begin{figure}[!t]
    \centering
    \includegraphics[width=0.9\columnwidth]{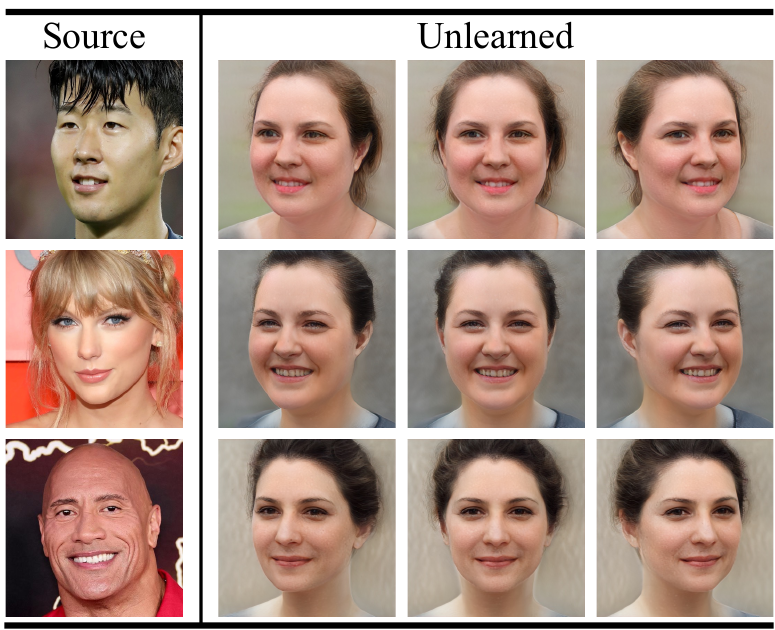}
    \caption{Given a single source image containing a specific identity, we remove that identity from the pre-trained 3D generative adversarial network (e.g., EG3D \cite{chan2022efficient}).
    Our method effectively unlearns identity even from in-the-wild images where the source image is absent in the pre-training dataset.}
    \label{fig:motivation}
\end{figure}

Recently, 2D or 3D Generative Adversarial Networks (GANs) \cite{karras2019style, karras2020analyzing, karras2021alias, chan2022efficient} pre-trained on large datasets, \eg, FFHQ \cite{karras2019style} or AFHQ \cite{choi2020starganv2}, have drawn substantial attention due to their remarkable generation performance and highly disentangled representation space.
However, their advancements have raised privacy concerns \cite{hayes2017logan}, especially regarding the potential misuse of generative models to represent and exploit individual identities.
For instance, deepfakes \cite{xu2023tall, yan2023ucf} can create very believable images or videos of people in made-up situations, causing major concerns about ethics and privacy.

To alleviate privacy issues in generative models, machine unlearning task has been actively studied.
Machine unlearning involves the process of selectively removing specific knowledge or erasing the influence of certain data from the training dataset of pre-trained models. It is beneficial especially when the data are harmful, private, or biased \cite{golatkar2020eternal,golatkar2020forgetting,mehta2022deep, peste2021ssse}.
Despite a focus on discriminative tasks in most machine unlearning research, a few studies have ventured into generative models, attempting to erase high-level concepts such as socially inappropriate content or artistic styles that present copyright challenges \cite{gandikota2023erasing, kumari2023ablating}.

Nevertheless, generative models still exhibit ongoing privacy issues.
Even if an identity of someone is not used in the pre-training of the generative models, it can be easily reconstructed in the pre-trained models via GAN inversion models \cite{richardson2021encoding, tov2021designing, roich2022pivotal, moon2022interestyle, yuan2023make, wang2022high}.
Furthermore, the reconstructed image can be manipulated or edited easily via image editing methods \cite{shen2020interpreting, patashnik2021styleclip, shen2020interfacegan}.
To prevent potential exploits of an identity, it is necessary to erase a certain identity from the pre-trained generative models.

To consider the above issue, we introduce an essential task of unlearning any identity from the pre-trained 2D or 3D GANs \cite{chan2022efficient, karras2020analyzing}, called \textit{\taskname{}}.
Unlike typical machine unlearning tasks, which focus on unlearning the training samples our \taskname{} task unlearns any identity on pre-trained GANs, even if it was not shown during the pre-training.
Our goal is to remove the whole identity associated with a given single image from the generator while minimally impacting the overall performance of the pre-trained model.

To achieve our goal, we propose a novel generative unlearning framework, \textbf{G}enerative \textbf{U}nlearning for Any \textbf{IDE}ntity, named \textbf{GUIDE}. \frameworkname{} replaces the source identity with an anonymous target identity, erasing the original identity effectively. To this end, we propose a new exploration method to determine an effective target latent code, called Un-Identifying Face On latent space (UFO). UFO utilizes the GAN inversion method \cite{yuan2023make} to embed the given identity into the source latent, and then decides the target latent using both the source and the average latent codes. We empirically find that the proposed UFO can identify the promising target to erase any given source identity robustly.



Given the source and target latent code, we update the generator to shift from the source identity to the target identity. To this end, we propose three novel loss functions: (\lowercase{\romannumeral1}) \textit{local unlearning loss}, (\lowercase{\romannumeral2}) \textit{adjacency-aware unlearning loss}, and (\lowercase{\romannumeral3}) \textit{global preservation loss}. (\lowercase{\romannumeral1}) guides our model directly shifting the source identity to the target identity. (\lowercase{\romannumeral2}) utilizes other latent codes adjacent to the source and target latent codes to effectively unlearn the entire identity from a single image. To minimize side effects from the unlearning process, (\lowercase{\romannumeral3}) additionally regularizes the generator to retain generation performance for latent codes relatively far from the source and target latent codes. Through comprehensive experiments on diverse identities, including \textit{Random}, \textit{InD}, and \textit{OOD}, we confirm that \frameworkname{} can successfully remove the identity of the source image from the pre-trained generative model, and shows qualitatively and quantitatively superior performances.

Our contributions can be summarized as follows:
\begin{itemize}
    \item For the first time, we propose a novel task, \taskname{}, which tackles machine unlearning in generative models in the aspect of privacy protection.
    In our task, we aim to prevent the pre-trained generative models from synthesizing the given identity by utilizing only a single image.
    \item For the effective and robust elimination of the identity, we propose a novel method - Un-Identifying Face On Latent Space (UFO).
    We configure the unlearning procedure by formulating how to represent and shift the identity in the latent space.
    We find that setting the extrapolated latent code between the source and average latent codes as an optimization target facilitates the unlearning procedure.
    \item We propose three loss functions - local unlearning loss, adjacency-aware unlearning loss, and global preservation loss to effectively unlearn the identity from the pre-trained model while less affecting the generation performance on other identity.
    \item We show that our proposed framework, \frameworkname{} achieves state-of-the-art performance both qualitatively and quantitatively, through extensive experiments. We demonstrate that \frameworkname{} can remove the specific identity successfully in the generative models while minimizing the negative effect on other identities. 
\end{itemize}
\section{Related Work}
\paragraph{Generative Models and Privacy Issue.}
In image synthesis field, GAN-based generative models have achieved remarkable performance not only in 2D \cite{Karrs2018progressive, karras2019style, karras2020analyzing, karras2021alias, ramesh2021zero, ramesh2022hierarchical, rombach2022high, seo2023lfs} but also in 3D \cite{chan2022efficient, sun2023next3d, zhao2022generative, wu2022anifacegan, zhang2023multi} domain.
The application of various image editing methods \cite{shen2020interfacegan, patashnik2021styleclip, yoo2019photorealistic} to strong generative models,
people can easily generate edited images of specific individuals and various artistic styles \cite{somepalli2023diffusion, shan2023glaze}, as well as extract copyrighted content \cite{carlini2023extracting} without the permission of the individual or the original creator.

Recently, with the rise of the importance of AI ethics, several works have addressed this issue \cite{gandikota2023erasing, kumari2023ablating, zhang2023forget, tiwary2023adapt, moon2023feature}.
ESD \cite{gandikota2023erasing} erases specific visual concepts from diffusion model by using negative guidance about the undesired concepts.
Kumari et al. \cite{kumari2023ablating} modifies the conditional distribution of the model a specific target concepts to match the anchor concept.
Forget-Me-Not \cite{zhang2023forget} fine-tunes U-Net to minimize each of the intermediate attention associated with the target concepts to remove.
Additional works in GANs \cite{tiwary2023adapt, moon2023feature} focus on unlearning specific features, \eg, “Bang”, “Hat” or “Beard” rather than forgetting specific identity.
The above methods primarily concentrate on the elimination of specific concepts or high-level features.
In other words, these cannot preclude models from generating specific individuals while maintaining the generation performance of realistic human faces.
Unlike the existing works, our work targets to unlearn only specific individuals without shifting the overall distribution of generated images.
\paragraph{Machine Unlearning.}
Machine unlearning aims to selectively forget specific acquired knowledge or diminish the impact of certain training data subsets on a trained model.
Since previous research \cite{fredrikson2015model,khosravy2022model,yuan2019adversarial} shows that machine learning models might accidentally share private information when faced with certain attacks or inputs, machine unlearning becomes crucial.

While previous machine unlearning is mainly focused on supervised learning tasks \cite{gupta2021adaptive, tarun2023fast, baumhauer2022machine, ginart2019making, golatkar2020eternal, golatkar2021mixed, chundawat2023zero, golatkar2020forgetting}, the interest in unlearning techniques within unsupervised learning, \ie, generative models, is growing \cite{kong2023data, tiwary2023adapt, moon2023feature}.
However, most of the existing methods need full dataset access for retraining, which is hard to acquire and computationally expensive \cite{gupta2021adaptive, tarun2023fast, baumhauer2022machine, ginart2019making}.
For example,  Kong and Chaudhuri \cite{kong2023data} utilizes data redaction and augmentation algorithms, which requires a full training dataset.
Despite the existence of a feature unlearning model \cite{moon2023feature} which does not need full dataset access, unlearning only an individual feature is not enough to forget a whole specific identity.
To this end, we propose an algorithm that enables forgetting the specific identity only with a single image.
Furthermore, our approach distinguishes itself from existing research by applying unlearning to unseen images, enabling the erasure of specific identities without prior exposure to those images.

\nocite{moon2023online}
\section{Method}
\begin{figure}[!t]
    \centering
    \includegraphics[width=0.45\textwidth]{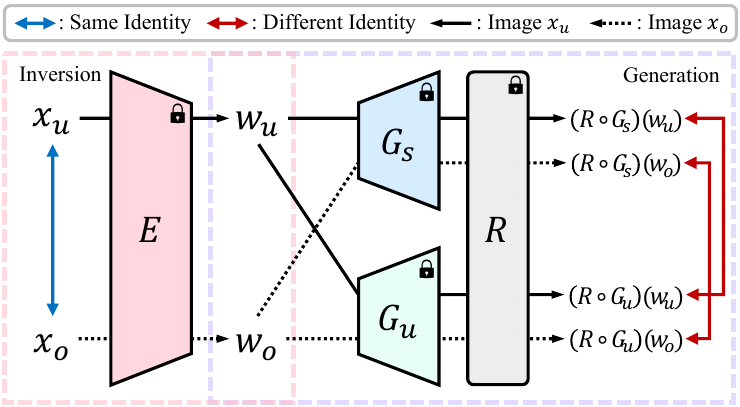}
    \caption{An illustration of \textit{\taskname{}}. 
    Upon GUIDE, the identity of the image generated from $w_u$, \ie, inversion of the source image $x_u$ by inversion network $E$, should exhibit a distinct identity when passed through the pre-trained generator $G_s$ compared to the unlearned generator $G_u$.
    Furthermore, other images $x_o$, not used in unlearning but sharing the same identity with $x_u$, also should vary an identity through GUIDE.
    }
    \label{fig:formulation}
\end{figure}

\begin{figure*}[!ht]
    \includegraphics[width=\textwidth]{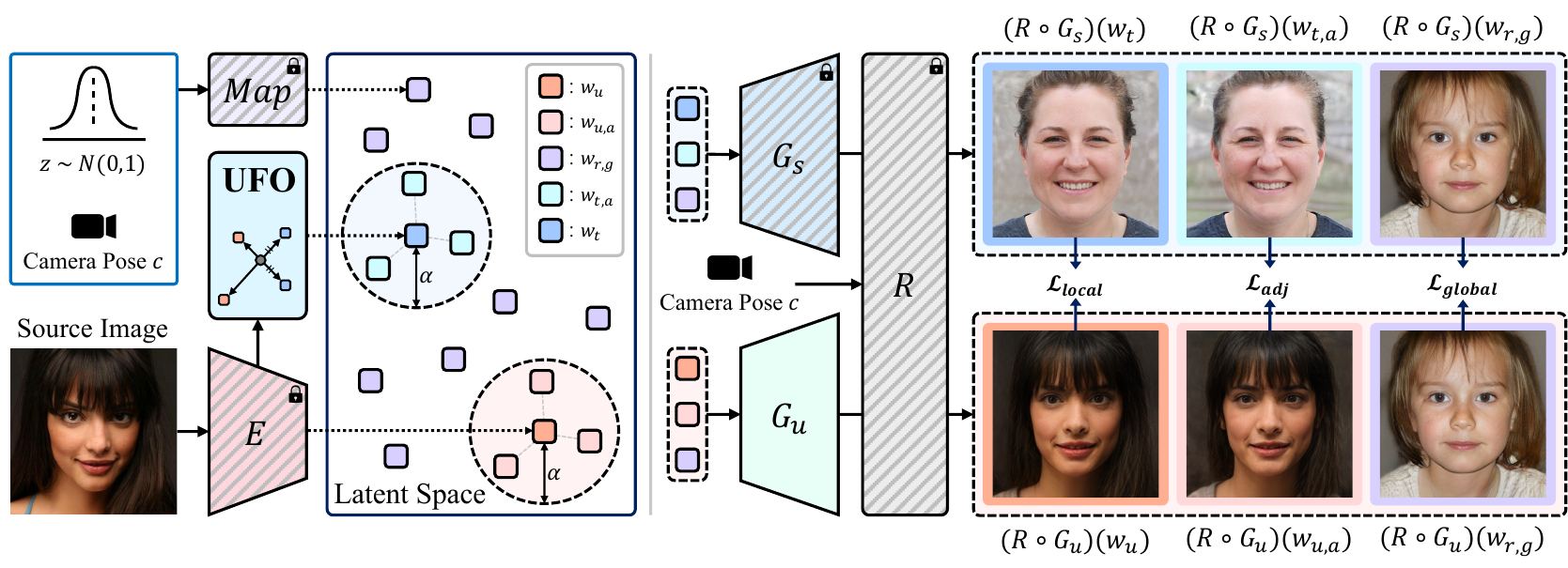}
    \caption{An overview of GUIDE. Starting with a source image, we employ a GAN inversion network $E$, specifically GOAE \cite{yuan2023make}, to embed this image into the latent space of a pre-trained generative model, namely EG3D \cite{chan2022efficient}, obtaining the source latent code $w_u$. The target latent code $w_t$ is designated through the UFO process.
    To facilitate identity removal in $w_u$, we shift its identity to match that of $w_t$ with our Latent Target Unlearning (LTU) process.
    Three loss functions of LTU are designed for this purpose: (i) The generator is optimized to produce an image from the source latent code, denoted as $(R \circ G_u)(w_u)$, that is similar to the image from the target latent code, represented as $(R \circ G_s)(w_t)$.
    (ii) To achieve unlearning across the entire identity, we consider latent codes near both the source and target latent codes, denoted as $w_{u,a}$ and $w_{t,a}$, respectively.
    (iii) To prevent model corruption during the unlearning process, we additionally sample latent codes from a random noise vector, represented as $w_{r,g}$, and optimize $G_u$ to preserve its generation ability on $w_{r,g}$.}
    \label{fig:structure}
\end{figure*}

\vspace{-0.1cm}
Firstly, in Section \ref{sec:problem_formulation}, as shown in Figure \ref{fig:formulation}, we introduce the problem we aim to address, named \taskname{}.
In Section \ref{method:ufo}, we introduce un-identifying face on latent space, which designates an appropriate target latent for unlearning.
Then, in Section \ref{method:ltu}, we introduce latent target unlearning, along with our proposed novel losses to unlearn the generator.
The total overview of our method is illustrated in Figure \ref{fig:structure}.

\subsection{Problem Formulation}
\label{sec:problem_formulation}
Given a set of images $\mathbf{x}$ depicting a specific identity, we randomly select a single source image $x_u\in\mathbf{x}$ as an exemplar of the identity.
Initially, using off-the-shelf inversion network \cite{yuan2023make} $E$ corresponding to the unconditional generator, \ie, EG3D \cite{chan2022efficient}, we embed $x_u$ to the source latent code $w_u$ in the latent space of EG3D:

\vspace{-0.5cm}
\begin{align}\label{eqn:inversion}
    w_u = E(x_u).
\end{align}
Since EG3D \cite{chan2022efficient} consists of the mapping network $Map(\cdot)$, StyleGAN2 \cite{karras2019style} backbone $G(\cdot)$ and the neural renderer with a super-resolution module $R(\cdot)$, we can denote the reconstructed image $\hat{x}$ from $w_u$ as following:

\vspace{-0.5cm}
\begin{align}
    \hat{x} = R(G(w_u); c),
\end{align}
where $c$ denotes camera poses.
For convenience, we omit the explicit notation of camera poses in this paper, \ie, $\hat{x} = (R \circ G)(w_u)$.
We target to derive an unlearned $G$, \ie, $G_u$, from the pre-trained EG3D generator $G$, \ie, $G_s$, while fixing $Map$ and $R$.
With proper unlearning, an image generated by unlearned EG3D using $w_u$, \ie, $\hat{x}_u = (R \circ G_u)(w_u)$ should have a distinct identity from the image generated by original EG3D using $w_u$, \ie, $(R \circ G_s)(w_u)$.

In our task formulation, two considerations are paramount.
First, we aim to eliminate the entire identity from the generator only utilizing a single image.
To validate this, we incorporate other multiple images $x_{o}\in\mathbf{x}$ for testing and its corresponding latent code $w_{o} = E(x_{o})$.
By utilizing $x_{o}$, we can verify whether $G_{u}$ has successfully unlearned the identity as a whole, rather than just unlearning the specific image $x_{u}$.
Second, we strive to maintain the generation performance of the pre-trained model.
To assess this, we sample multiple images from fixed latent codes using both the unlearned and pre-trained generators.
We then estimate the distribution shift between the images generated before and after the unlearning process.

\begin{figure}[!t]
    \centering
    \includegraphics[width=0.8\columnwidth]{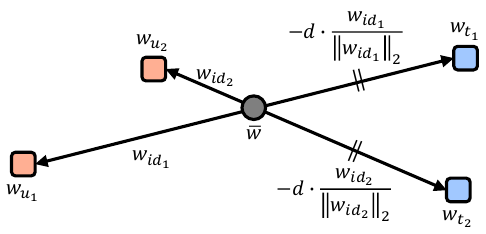}
    \caption{An illustration of Un-identifying Face On Latent Space (UFO). We define the identity of the source latent code by subtract it from the average latent code. We set the target latent code for our unlearning process by measuring an extrapolation between the source and average latent code with a fixed distance $d$.}
    \vspace{-0.5cm}
    \label{fig:ufo}
\end{figure}
\subsection{Un-Identifying Face On Latent Space}
\label{method:ufo}

The successfully unlearned model should not generate the image with the identity of $\mathbf{x}$, even when $w_u$ is used as a latent.
Consequently, we initiate our approach to manipulate $\hat{x}_u$ to be another image rather than the image with identity in $\mathbf{x}$ by unlearning $G$.
To design the objective function for unlearning, we first need to establish the objective for unlearning, which involves defining the target image $\hat{x}_t$, derived from $w_t$, which the unlearned image $\hat{x}_u$, derived from $w_u$, should mimic after unlearning.
While there exist various options for setting $\hat{x}_t$, \eg, a random face or even a non-human image, we choose the mean face generated by the mean latent $\Bar{w}$ of $Map$, \ie, $(R \circ G_s)(\Bar{w})$, as $\hat{x}_t$.
This setting is inspired by inversion methods \cite{alaluf2021restyle, moon2022interestyle} which apply the identity to the mean face through inversion stages.
Since our goal is to de-identify the image, we argue that it is intuitive and reasonable to send the image back to the mean face, which is opposite to the inversion.

However, this simple approach might be problematic when $w_u$ is close to $\Bar{w}$, where the unlearned image might still resemble the original identity.
To this end, we propose a novel method named Un-identifying Face On latent space (UFO), which can set $\hat{x}_t$ robustly regardless of the distance between $w_{u}$ and $\Bar{w}$, as shown in Figure \ref{fig:ufo}.
The following are the processes of UFO:
First, in the \textit{de}-identification process, we extract the identity latent $w_{id} = w_{u} - \Bar{w}$ by subtracting $\Bar{w}$ from $w_{u}$.
Then we propose a process of \textit{en}-identification, setting the target in the opposite direction of the existing $w_{id}$ to foster a more pronounced change in identity, creating an entirely new identity.
In other words, we chose $\Bar{w}$ as the stopover point and established the final target point at the extrapolation of $w_u$ and $\Bar{w}$.
The finalized target point can be expressed as:

\vspace{-0.5cm}
\begin{align}
    w_t = \Bar{w} -  d \cdot \frac{w_{i d}}{\left\|w_{i d}\right\|_2},
\end{align}
where $d$ is defined as the distance that balances un-identification with preservation of the source distribution, as determined by our empirical results.
Finally, we can set $\hat{x}_t=(R \circ G_s)(w_t)$.
We empirically demonstrate that whether the given identity is in close proximity to the average latent code within the latent space or not, UFO sets the desirable target latent code for the generative unlearning.

\subsection{Latent Target Unlearning}
\label{method:ltu}

After setting $\hat{x}_t$, we need to robustly design the unlearning process that can effectively make $\hat{x}_u$ to be similar to $\hat{x}_t$, while maintaining the generation performance of $G_u$.
We coin this process as Latent Target Unlearning (LTU), which targets to unlearn images from the specific latent while keeping generation results from other latent codes.
LTU utilizes the following three losses to achieve this goal:


\paragraph{Local Unlearning Loss.}
To force $\hat{x}_u$ to be $\hat{x}_t$, we use the widely-used reconstruction losses, \ie, Euclidean loss $\mathcal{L}_{2}$, perceptual loss  $\mathcal{L}_{per}$ \cite{zhang2018unreasonable}, and identity loss  $\mathcal{L}_{id}$ \cite{deng2019arcface}.
Using $\mathcal{L}_{recon}$, we compare the tri-plane features $F_u = G_u(w_u)$ and $F_t = G_s(w_t)$, derived from source and target latent codes, respectively.
The local unlearning loss is defined as:
\vspace{-0.3cm}
\begin{align}
\label{eq:local}
    \begin{split}
        \mathcal{L}_{local}(\hat{x}_u, \hat{x}_t) &=\lambda_{L2}\mathcal{L}_{L2}(F_u, F_t) \\
        &+ \lambda_{per}\mathcal{L}_{per}(\hat{x}_u, \hat{x}_t)
        + \lambda_{id}\mathcal{L}_{id}(\hat{x}_u, \hat{x}_t).
    \end{split}   
\end{align}
By adopting $\mathcal{L}_{local}$, we can successfully un-identify the given source identity in $\hat{x}_t$.

\begin{figure}[!t]
    \centering
    \includegraphics[width=\columnwidth]{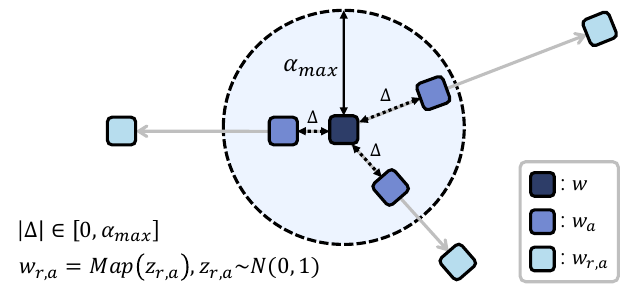}
    \caption{An illustration of determining latent codes near a latent code $w$ in adjacency-aware unlearning loss. We first sample a latent code $w_{r,a}$ which is derived from a random noise vector $z_{r,a}$ via the mapping network $Map(\cdot)$, \ie $w_{r,a}=Map(z_{r,a})$. Next, we compute the direction between $w$ and $w_{r,a}$, and we scale it to fall within range between 0 and $\alpha_{max}$. This yields the distance vector $\Delta$ to compute the adjacent latent code $w_a=w+\Delta$.}
    \vspace{-0.3cm}
    \label{fig:adj}
\end{figure}
\vspace{-0.3cm}
\paragraph{Adjacency-Aware Unlearning Loss.}
The above equation considers only one pair of source and target latent codes.
However, images of a similar identity to the source identity can be obtained by introducing marginal perturbations to the latent code.
For the successful unlearning of the given identity, we need to consider the neighborhood of both the source and the target latent codes.
Consequently, as shown in Figure \ref{fig:adj}, we sample $N_a$ latent codes in the vicinity of the $w_u$.
Specifically, with the scale $\alpha^i$ sampled from the uniform distribution with hyperparameter $\alpha_{max}$, \ie $\alpha^i \sim \mathcal{U}(0, \alpha_{max})$, we define the distances $\Delta$ to compute the adjacent latent codes as:
\begin{align}
    \Delta = \{\alpha^i \cdot \frac{w^i_{r,a} - w_u}{\Vert w^i_{r,a} - w_u \Vert_2}\}^{N_a}_{i=1},
\end{align}
where $w^i_{r,a}$ is a latent code sampled from the random noise vector $z^i_{r,a}$.
Using these distances $\Delta$, we can compute $N_a$ latent codes for both the source and the target latent codes.
Similar to the local unlearning loss, we optimize the generated tri-plane features and images from $w^i_{u,a} = w_u + \Delta^i$ and $w^i_{t,a} = w_t + \Delta^i$ to be similar:

\begin{align}
    &\hat{x}^i_{u,a} = R(F^i_{u,a}), \hat{x}^i_{t,a} = R(F^i_{u,a}),\\
    &\mathcal{L}_{adj}(w_u, w_t) = \cfrac{1}{N_a}\sum\limits^{N_a}_{i=1}\mathcal{L}_{local}(\hat{x}^i_{u,a}, \hat{x}^i_{t,a}),
\end{align}
where $F^i_{u,a} = G_u(w^i_{u,a}), F^i_{t,a} = G_s(w^i_{t,a})$ denotes for tri-plane features, and $\mathcal{L}_{local}$ in Equation \ref{eq:local}.
From $\mathcal{L}_{adj}$, we can further consider possible variations of the source identity.

\paragraph{Global Preservation Loss.}
While the local unlearning loss and adjacency-aware unlearning loss mentioned above facilitate the removal of the source identity, we propose a global preservation loss to mitigate side effects arising from these unlearning loss functions.
In the global preservation loss, we constrain the generator to maintain generation performance for latent codes that are relatively distant from both the source and target latent codes.

To be precise, we sample $N_g$ latent codes $\{w^i_{r,g}\}^{N_g}_{i=1}$ from random noise vectors $\{z^i_{r,g}\}^{N_g}_{i=1}$.
We ensure that these do not overlap with the adjacent latent codes used in the adjacency-aware unlearning loss.
Unlike the unlearning loss functions, we find that adopting only $L_{per}$ achieves a balanced performance between identity shift and model preservation.
The global preservation loss is computed as:

\vspace{-0.3cm}
\begin{align}
    \begin{split}
    \hat{x}^i_{u,g} &= (R \circ G_u)(w^i_{r,g}), \\
    \hat{x}^i_{s,g} &= (R \circ G_s)(w^i_{r,g}), \\
    \mathcal{L}_{global}(G_u, G_s) &= \cfrac{1}{N_g}\sum\limits^{N_g}_{i=1}\mathcal{L}_{per}(\hat{x}^i_{u,g}, \hat{x}^i_{s,g}).
    \end{split}
\end{align}
In summary, our final objective is:
\begin{align}
\begin{split}
    \mathcal{L}_{total} = \mathcal{L}_{local} + \lambda_{adj}\mathcal{L}_{adj} + \lambda_{global}\mathcal{L}_{global}. 
\end{split}
\end{align}

\section{Experiments}
\subsection{Experimental Setup}
\label{sec:experimental_setup}
\paragraph{Baseline.} Since we propose \taskname{} task for the first time, to evaluate the effectiveness of \frameworkname{}, we constructed a simple baseline.
In the baseline, we used the target latent code as the average one for the unlearning.
During the unlearning, we updated the pre-trained generator using $\mathcal{L}_{local}$ as described in Equation \ref{eq:local}.
\begin{figure*}[!t]
    \centering
    \includegraphics[width=\textwidth]{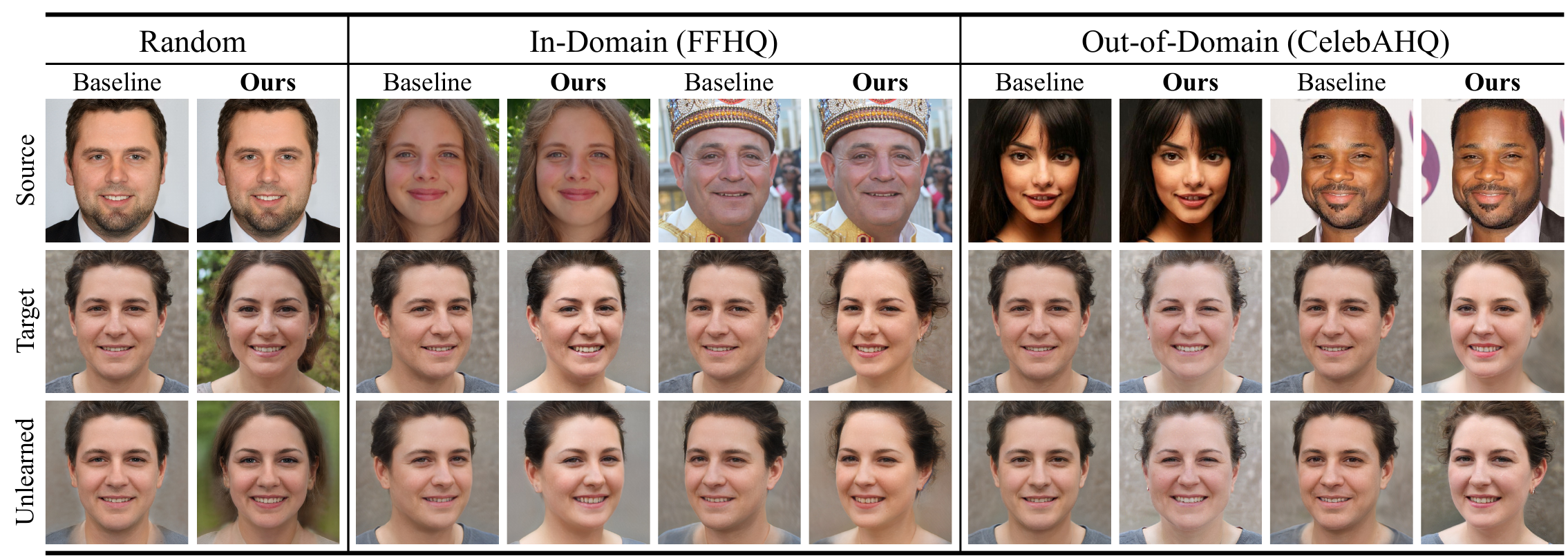}
    \caption{Qualitative results of \frameworkname{} and the baseline in \taskname{} task. For the given source image each (the first row), \frameworkname{} and the baseline tried to erase the identity in the pre-trained generator. The images in the second and third row are the target and unlearned images, respectively.}
    \label{fig:single-image}
\end{figure*}

\paragraph{Implementation Details.}
We built \frameworkname{} based on the 3D generative adversarial network \cite{chan2022efficient} pre-trained on FFHQ dataset \cite{karras2019style}.
We used GOAE \cite{yuan2023make} as a GAN inversion network to obtain the latent code from an image.
The image resolution we used in our experiments is 512x512 with a rendering resolution of 128x128.
We used Adam optimizer \cite{kingma2015adam} with a learning rate of $10^{-4}$ in the unlearning procedure.
The hyperparameters used in the experiments were: $d=30$, $\alpha_{max}=15$, $\lambda_{L2}=10^{-2}$, $\lambda_{per}=1$, $\lambda_{id}=10^{-1}$, $N_a=N_g=2$, and $\lambda_{adj}=\lambda_{global}=1$.
\paragraph{Dataset and Scenarios.}
We evaluated \frameworkname{} in three scenarios: \textit{Random}, where we set an unlearning target image from a randomly sampled noise vector; \textit{InD} (\textit{in-domain}), where we sampled an image from the FFHQ dataset \cite{karras2019style} used for pre-training; and \textit{OOD} (\textit{out-of-domain}), where the unlearning target image was sampled from the CelebAHQ dataset \cite{Karrs2018progressive}. 
For \textit{InD} and \textit{OOD}, we used the GAN inversion network to obtain corresponding latent codes. 
For \textit{OOD} scenario, we also conducted \textit{multi-image} test since there were multiple images with a same identity in CelebAHQ. 
On the other hand, we performed only \textit{single-image} test in the \textit{Random} and \textit{InD} scenarios.

\paragraph{Evaluation Metrics.}
We evaluated \frameworkname{} on two key aspects.
Firstly, we estimated the efficacy of our approach in preventing the generator from producing images similar to the unlearning target.
We quantitatively measured similarity of identities (ID) using face recognition network, CurricularFace \cite{huang2020curricularface}, between images generated from the same latent codes before and after unlearning.
Moreover, we utilized ID\textsubscript{others} to estimate the erasure of a identity from images which are unseen during training but containing the same identity of the source image.
Secondly, we assessed whether our method preserves overall generation performance using the Fr\'echet Inception distance (FID) score \cite{heusel2017gans}.
Different from the existing usages, we utilized two variants of FID. First, we evaluated the distribution shift of generated images the pre-trained generator and the unlearned generator via FID\textsubscript{pre}.
Furthermore, we measured the distribution shift with respect to the real FFHQ images, which we denoted as $\Delta$FID\textsubscript{real}.

\subsection{Main Experiment}
\begin{figure}[!t]
    \centering
    \includegraphics[width=0.86\columnwidth]{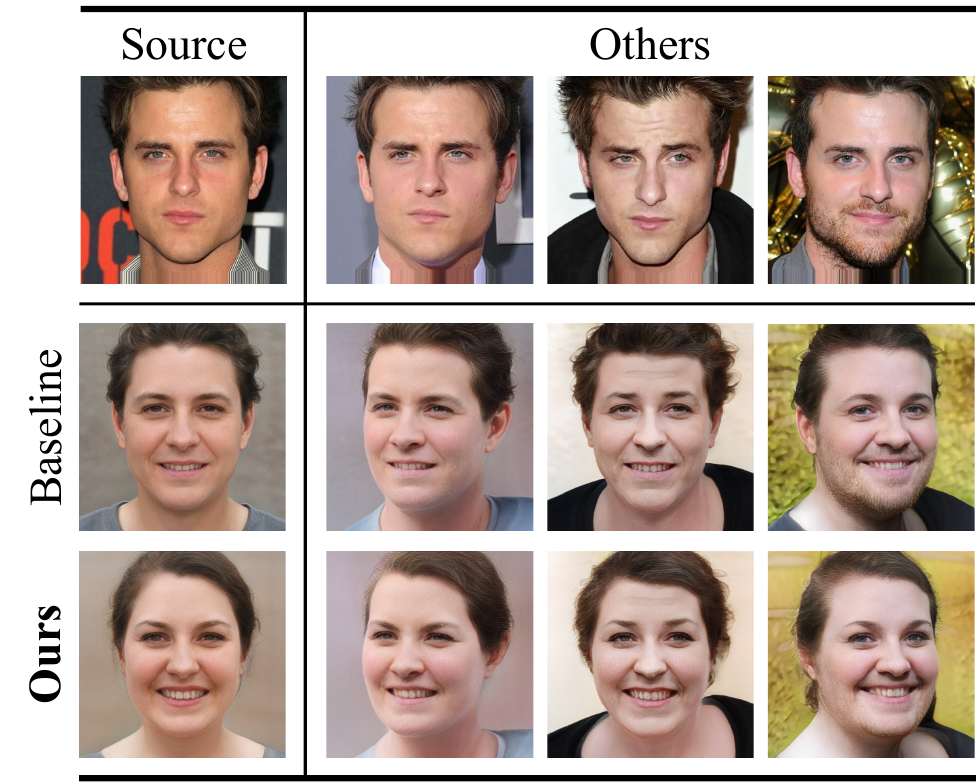}
    \caption{Qualitative results of \frameworkname{} and the baseline on a multi-image test using CelebAHQ dataset.
    We additionally utilized images that are unseen during unlearning, to show how thoroughly erase the given identity.}
    \label{fig:multi-image}
    \vspace{-0.4cm}
\end{figure}
\subsubsection{Qualitative Results}
\vspace{-0.15cm}
We conducted a comparative analysis of \frameworkname{} against the baseline in the \taskname{} task.
Initiating from the provided source image, we aimed to eliminate the identity within the pre-trained generator, as illustrated in Figure \ref{fig:single-image}.
We presented the resulting unlearned image, along with the target image optimized in our loss functions.
Notably, \frameworkname{} effectively erases identities whether synthetic, presented during pre-training, or unseen.

\begin{figure}[!t]
    \centering
    \includegraphics[width=\columnwidth]{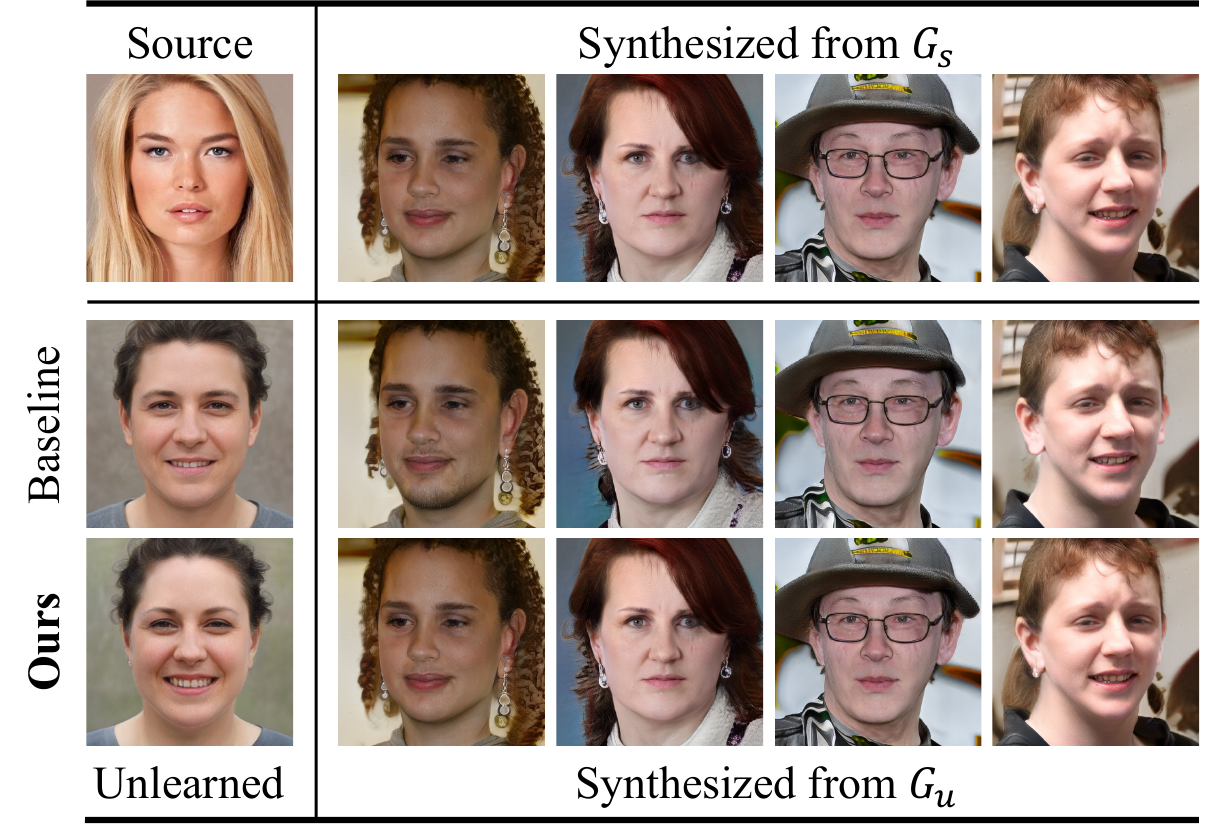}
    \caption{Qualitative comparison between \frameworkname{} and the baseline on the preservation of the generation quality of other identities. GUIDE generates images almost identical to those synthesized by $G_s$, whereas the baseline often results in noticeable changes, \eg, beard shape, hairstyle change, hat.}
    \vspace{-0.3cm}
    \label{fig:preserve}
\end{figure}
\begin{table*}[!h]
\centering
\resizebox{\textwidth}{!}{%
\begin{tabular}{l|ccc|ccc|ccc}
\toprule[1.1pt]
\multirow{2}{*}{Methods} & \multicolumn{3}{c|}{Random} & \multicolumn{3}{c|}{In-Domain (FFHQ)} & \multicolumn{3}{c}{Out-of-Domain (CelebAHQ)} \\ \cline{2-4}\cline{5-7}\cline{8-10}
 &
  ID ($\downarrow$) &
  FID\textsubscript{pre} ($\downarrow$) &
  $\Delta$FID\textsubscript{real} ($\downarrow$) &
  ID ($\downarrow$) &
  FID\textsubscript{pre} ($\downarrow$) &
  $\Delta$FID\textsubscript{real} ($\downarrow$) &
  ID ($\downarrow$) &
  FID\textsubscript{pre} ($\downarrow$) &
  $\Delta$FID\textsubscript{real} ($\downarrow$) \\ \hline
Baseline                & 0.19 \scriptsize{$\pm$ 0.09}   & 11.73 \scriptsize{$\pm$ 2.74}   & 7.46 \scriptsize{$\pm$ 2.20} 
                        & 0.16 \scriptsize{$\pm$ 0.07}   & 9.00 \scriptsize{$\pm$ 1.15}   & 4.15 \scriptsize{$\pm$ 1.18}  
                        & 0.12 \scriptsize{$\pm$ 0.06}   & 9.52 \scriptsize{$\pm$ 1.53}   & 4.75 \scriptsize{$\pm$ 0.89}    \\ \hdashline
+ extrapolated $w_t$    & \textbf{0.12 \scriptsize{$\pm$ 0.06}}   & 14.28 \scriptsize{$\pm$ 3.34}   & 9.63 \scriptsize{$\pm$ 2.53} 
                        & 0.05 \scriptsize{$\pm$ 0.06}   & 12.78 \scriptsize{$\pm$ 1.82}   & 6.76 \scriptsize{$\pm$ 1.41}  
                        & 0.02 \scriptsize{$\pm$ 0.05}   & 13.02 \scriptsize{$\pm$ 3.20}   & 7.31 \scriptsize{$\pm$ 1.98}      \\
+ $\mathcal{L}_{adj}$   & 0.14 \scriptsize{$\pm$ 0.07}   & 19.65 \scriptsize{$\pm$ 4.90}   & 13.94 \scriptsize{$\pm$ 3.59} 
                        & \textbf{0.04 \scriptsize{$\pm$ 0.06}}   & 13.53 \scriptsize{$\pm$ 2.08}   & 7.35 \scriptsize{$\pm$ 1.70}  
                        & \textbf{0.01 \scriptsize{$\pm$ 0.05}}   & 13.63 \scriptsize{$\pm$ 3.52}   & 7.83 \scriptsize{$\pm$ 2.19}      \\ \hdashline
+ $\mathcal{L}_{global}$ (\textbf{\frameworkname{}}) & 0.14 \scriptsize{$\pm$ 0.06}   & \textbf{10.80 \scriptsize{$\pm$ 2.70}}   & \textbf{6.64 \scriptsize{$\pm$ 1.60}} 
                        & 0.06 \scriptsize{$\pm$ 0.06}   & \textbf{8.00 \scriptsize{$\pm$ 1.20}}   & \textbf{3.05 \scriptsize{$\pm$ 0.81}}  
                        & 0.03 \scriptsize{$\pm$ 0.05}   & \textbf{7.88 \scriptsize{$\pm$ 1.96}}   & \textbf{3.34 \scriptsize{$\pm$ 1.10}}      \\ \bottomrule[1.1pt]
\end{tabular}%
}
\caption{Quantitative results of \frameworkname{} and the baseline in the \taskname{} task, tested in a single-image setting using one image per identity. Starting from the baseline, we gradually introduced components of \frameworkname{}.}
\label{tab:single_image}

\end{table*}
To evaluate the thoroughness of identity removal, we performed a multi-image test using identities from the CelebAHQ dataset.
This test involved assessing the ID similarity not only for the unlearned image derived from the source image but also for other images sharing the same identity.
As shown in Figure \ref{fig:multi-image}, \frameworkname{} showed superior generalization for unseen images compared to the baseline.
This improvement is attributed to the adjacency-aware unlearning, which facilitated the unlearning process not just for the given images but also for their neighborhood.

In Figure \ref{fig:preserve}, we conducted an experiment to assess the effect of the unlearning process on other identities.
While the baseline had a significant impact on other identities through the unlearning, \frameworkname{} showed a relatively lesser effect.
We attribute this to the global preservation loss, which constrained the distribution shift on other latent codes.

\vspace{-0.3cm}
\subsubsection{Quantitative Results}
In Table \ref{tab:single_image}, we compared \frameworkname{} to the baseline by gradually applying the components of \frameworkname{}.
By configuring $w_t$ through extrapolation, we achieved performance improvements in ID similarity across three scenarios.
Notably, we observed a significant difference in ID similarities in the random scenario, indicating that in cases where a latent code was close to $\Bar{w}$, \ie, as in the random scenario, there was insufficient removal of identity.
The effectiveness of employing an extrapolation between the source latent code and the average latent code was evident in such instances.

The adjacency-aware unlearning loss further enhanced the unlearning an identity.
This loss was designed to cover the vicinity of the source latent code, thereby promoting unlearning on the source latent code itself.
Finally, the application of the global preservation loss effectively reduced the estimated distribution shift using FID\textsubscript{pre} and $\Delta$FID\textsubscript{real}.

Moreover, we conducted a multi-image test in an \textit{OOD} scenario.
In this particular experiment, we introduced additional metric - ID\textsubscript{others} aimed at quantifying ID similarities for the unseen images associated with the source identity.
As presented in Table \ref{tab:multi_image}, the introduction of the adjacency-aware unlearning loss resulted in a remarkable improvement in ID\textsubscript{others}, emphasizing the effectiveness of this unlearning approach for handling unseen images.

\subsection{Ablation Study}
\begin{table}[!t]
\resizebox{\columnwidth}{!}{%
\begin{tabular}{l|ccccc}
\toprule[1.1pt]
Methods &
  ID ($\downarrow$) &
  ID\textsubscript{others} ($\downarrow$) &
  FID\textsubscript{pre} ($\downarrow$) &
  $\Delta$FID\textsubscript{real} ($\downarrow$) \\ \hline
Baseline                    & 0.12 \scriptsize{$\pm$ 0.06}  & 0.28 \scriptsize{$\pm$ 0.08}
                            & 9.52 \scriptsize{$\pm$ 1.53}  & 4.75 \scriptsize{$\pm$ 0.89} \\ \hdashline
+ extrapolated $w_t$        & 0.02 \scriptsize{$\pm$ 0.05}  & 0.15 \scriptsize{$\pm$ 0.07} 
                            & 13.02 \scriptsize{$\pm$ 3.20} & 7.31 \scriptsize{$\pm$ 1.98} \\
+ $\mathcal{L}_{adj}$       & \textbf{0.01 \scriptsize{$\pm$ 0.05}}  & \textbf{0.14 \scriptsize{$\pm$ 0.07}}  
                            & 13.63 \scriptsize{$\pm$ 3.52} & 7.83 \scriptsize{$\pm$ 2.19} \\ \hdashline
+ $\mathcal{L}_{global}$ (\textbf{\frameworkname{}}) & 0.03 \scriptsize{$\pm$ 0.05} & 0.17 \scriptsize{$\pm$ 0.08}  
                                                & \textbf{7.88 \scriptsize{$\pm$ 1.96}} & \textbf{3.34 \scriptsize{$\pm$ 1.10}} \\ \bottomrule[1.1pt]
\end{tabular}%
}
 \caption{Quantitative results of \frameworkname{} and the baseline in the generative identity unlearning in a multi-image setting, \ie, using a single image for unlearning and the other images for testing. We used CelebAHQ dataset for this test.}
\vspace{-0.4cm}
\label{tab:multi_image}
\end{table}
\paragraph{Effect of $d$ in Determination of $w_t$.}
\begin{figure}[!t]
    \centering
    \includegraphics[width=0.45\textwidth]{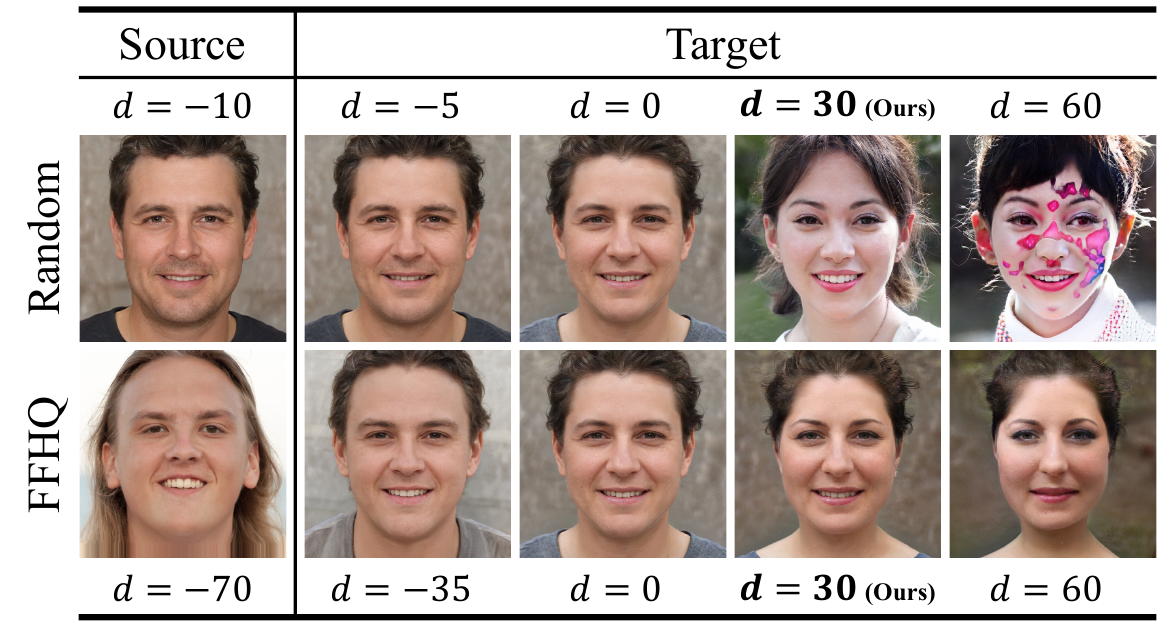}
    \caption{Ablation study to figure out the effectiveness of $d$. We visualized target images corresponding to each source image with different values of $d$. The target images were generated using target latent codes derived from interpolated latent codes, the average latent code ($d=0$), or extrapolated latent codes ($d>0$). Interpolation and extrapolation were carried out between the source and the average latent code. In the case of interpolation, the center between the source and the average latent code was computed.}
    \label{fig:ablation_d}
\end{figure}

We conducted an ablation study by comparing target images derived from varied values of $d$.
Setting $d$ to 0 denotes utilizing the $\Bar{w}$ as $w_{t}$ in the unlearning process.
For $d < 0$, we designated $w_{t}$ as an interpolated latent code between the $w_{s}$ and the $\Bar{w}$.
Conversely, for $d > 0$, we employed an extrapolated $w_{t}$, as detailed in Section \ref{method:ufo}.
As illustrated in Figure \ref{fig:ablation_d}, when $d < 0$, the target image closely aligns with the given source images.
However, as $d$ deviates from 0, the quality of the target image rapidly deteriorates, resulting in a pronounced collapse in the distribution of the pre-trained generator.
Consequently, the effectiveness of unlearning with such target images diminishes in removing identity from the source images.
Setting $d$ to 0 might suggest the use of $\Bar{w}$ as an effective target for erasing identity.
However, our ablation studies indicate that when the source image closely aligns with $\Bar{w}$, the unlearning procedure fails to thoroughly eliminate the identity.
Conversely, when $d > 0$, $w_{t}$ contains a distinct identity compared to the source image while maintaining a consistent distance from $\Bar{w}$.
Among the instances where $d > 0$, our ablation studies reveal that setting $d = 30$ achieves a balanced performance between effective unlearning and preservation of the generation performance of the pre-trained model.
\vspace{-0.3cm}

\paragraph{Effect of $\alpha_{max}$ in $\mathcal{L}_{adj}$.}
\begin{table}[!t]
\centering
\resizebox{0.6\columnwidth}{!}{%
\begin{tabular}{c|ccccc}
\toprule[1.1pt]
$\alpha_{max}$ & ID ($\downarrow$) & ID\textsubscript{others} ($\downarrow$) \\ \hline
0  & 0.1205 \scriptsize{$\pm$ 0.0603}  & 0.2754 \scriptsize{$\pm$ 0.0791}  \\
10 & 0.0892 \scriptsize{$\pm$ 0.0620}  & 0.2123 \scriptsize{$\pm$ 0.0762}  \\
\textbf{15} & \textbf{0.0878 \scriptsize{$\pm$ 0.0375}}  & \textbf{0.2094 \scriptsize{$\pm$ 0.0692}}  \\
20 & 0.0900 \scriptsize{$\pm$ 0.0538}  & 0.2105 \scriptsize{$\pm$ 0.0924}  \\
30 & 0.0926 \scriptsize{$\pm$ 0.0561}  & 0.2111 \scriptsize{$\pm$ 0.0653}  \\ \bottomrule[1.1pt]
\end{tabular}%
}
\caption{Ablation study to figure out the effectiveness of $\mathcal{L}_{adj}$ and $\alpha_{max}$. We compared the performance based on how successfully the given identity was erased, using ID and ID\textsubscript{others} metric. The row where $\alpha_{max}=0$ denotes the baseline. We used CelebAHQ dataset in this experiment.}
\vspace{-0.3cm}
\label{tab:ablation_alpha}
\end{table}
In Table \ref{tab:ablation_alpha}, we scrutinized the effectiveness of the adjacency-aware unlearning loss.
To ensure a fair comparison, we employed $\Bar{w}$ as $w_{t}$ in this experiment, and we used $\mathcal{L}_{local}$ and $\mathcal{L}_{adj}$ in the unlearning procedure.
Rows corresponding to $\alpha_{max}=0$ represent experimental results without the incorporation of $\mathcal{L}_{adj}$ in the unlearning procedure.
The introduction of $\mathcal{L}_{adj}$ resulted in consistent performance gains in ID\textsubscript{others}. 
This observation highlights the efficacy of considering not only the pair of source and target latent codes but also their surroundings for unlearning the entire identity.
\begin{table}[!t]
\centering
\resizebox{0.7\columnwidth}{!}{%
\begin{tabular}{cc|cc}
\toprule[1.1pt]
$\mathcal{L}_{local}$ &  $\mathcal{L}_{global}$ & FID\textsubscript{pre} ($\downarrow$) & $\Delta$FID\textsubscript{real} ($\downarrow$) \\ \hline
\checkmark &      & 9.52 \scriptsize{$\pm$ 1.53}  & 4.75 \scriptsize{$\pm$ 0.89}  \\
\checkmark &    \checkmark & \textbf{4.63 \scriptsize{$\pm$ 0.43}}  & \textbf{1.48 \scriptsize{$\pm$ 0.29}}  \\ \bottomrule[1.1pt]
\end{tabular}%
}
\caption{Ablation study to figure our the effectiveness of $\mathcal{L}_{global}$. We compared how preserved the performance of the pre-trained model through the unlearning process, via FID\textsubscript{pre} and $\Delta$FID\textsubscript{real}. We used CelebAHQ dataset in this experiment.}
\label{tab:ablation_global}
\end{table}
\paragraph{Effect of $\mathcal{L}_{global}$.}
To assess the effectiveness of the global preservation loss, a similar experiment was conducted as the previous experiment, \ie, setting $\Bar{w}$ as $w_{t}$.
The results are presented in Table \ref{tab:ablation_global}. The application of $\mathcal{L}_{global}$ demonstrated consistent performance improvements in both FID\textsubscript{pre} and $\Delta$FID\textsubscript{real}.
This suggests that imposing constraints on the generator to maintain its generation performance in latent codes distant from our primary focus is effective in reducing distribution shifts in generative models.

\section{Conclusion}
In this paper, we introduced a novel task, referred to as \taskname{}, designed to address privacy concerns in pre-trained generative adversarial networks.
This task requires thoroughly removing the identity of a single source image from the pre-trained generator.
To achieve this, we proposed a new framework, GUIDE (Generative Unlearning for any IDEntity). To unlearn the single identity, we first defined the target latent code via extrapolation, moving away from the average latent by the pre-defined distance in the direction from the source to the average latent. Using this, our GUIDE successfully unlearned the given identity via Latent Target Unlearning (LTU), which optimized the pre-trained model to preserve the overall generative ability but not to generate the same identity within the local space. Experimental results demonstrated the effectiveness of GUIDE with promising outcomes. We anticipate that our work will be widely applied in research or the industry field, providing users with a sense of freedom from privacy concerns through identity removal.
\vspace{-0.15cm}
\section*{Acknowledgement}
\vspace{-0.1cm}
This work was supported by MSIT (Ministry of Science and ICT), Korea, under the ITRC (Information Technology Research Center) support program (IITP-2024-RS-2023-00258649) supervised by the IITP (Institute for Information \& Communications Technology Planning \& Evaluation), and in part by NRF-2023S1A5A2A21083590, and in part by the IITP grant funded by the Korea Government (MSIT) (Artificial Intelligence Innovation Hub) under Grant 2021-0-02068, and by the IITP grant funded by the Korea government (MSIT) (No.RS-2022-00155911, Artificial Intelligence Convergence Innovation Human Resources Development (Kyung Hee University)).
\vspace{-1cm}
\appendix
\maketitlesupplementary
\begin{figure}[!t]
    \centering
    \includegraphics[width=\columnwidth]{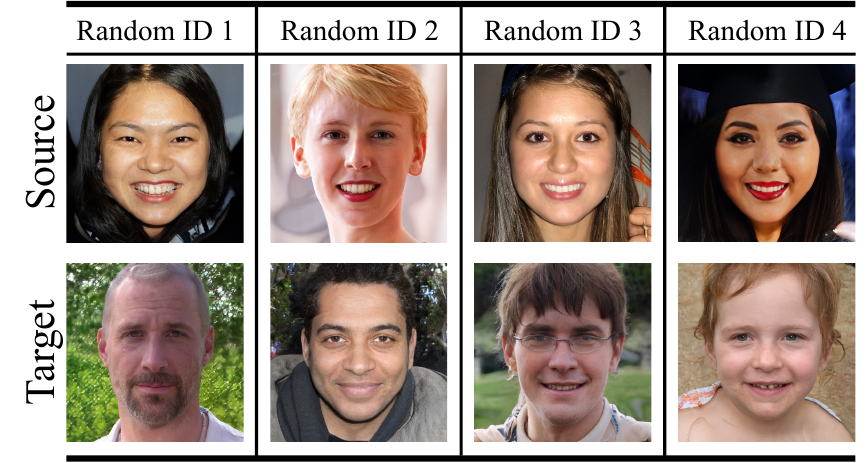}
    \caption{Illustration of diverse target images from randomly generated source images.}
    \label{fig:diverse}
\end{figure}
\section{Additional Experiments}
\subsection{Target Images from Diverse Source Images}
In Figure \ref{fig:diverse}, we visualized target images corresponding to the diverse source images.
In this experiment, we used randomly sampled $w_u$ to find the corresponding $w_t$.
Instead of the average latent code, by setting an extrapolated latent code as a target, we can obtain the effective and diverse target images for unlearning procedure.


\begin{figure}[!t]
    \centering
    \includegraphics[width=0.8\columnwidth]{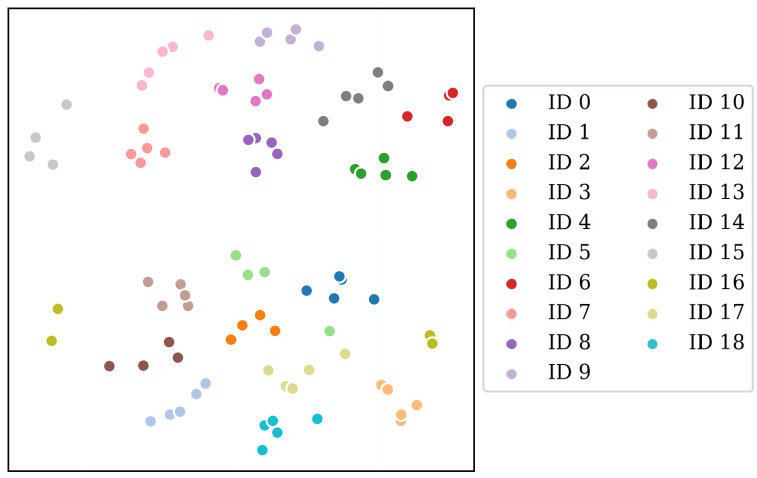}
    \caption{The relationship between the images and their identities in the latent space with t-SNE \cite{maaten2008visualizing}. Points of the same color denote the same identity. We used 5 images per identity from CelebAHQ dataset.}
    \vspace{-0.3cm}
    \label{fig:tsne_celebahq}
\end{figure}
\subsection{Distribution of Identities within CelebAHQ}
In designing \frameworkname{}, we assume that images sharing the same identity tend to cluster together.
Consequently, considering the proximity of latent codes aids in a more comprehensive erasure of identity.
Figure \ref{fig:tsne_celebahq} illustrates the relationship between images and their respective identities, utilizing 5 images per identity.
Our observation reveals a close grouping of images from the same identity in the latent space.
This finding aligns with the effectiveness of the $\mathcal{L}_{adj}$ proposed in Section 3.3 of our main paper.

\begin{figure}[!t]
    \centering
    \includegraphics[width=\columnwidth]{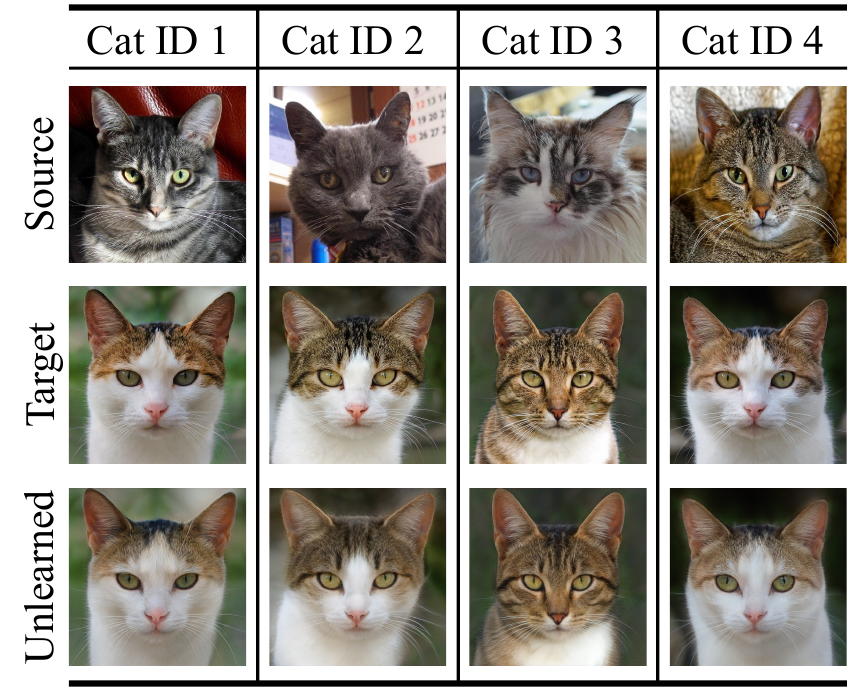}
    \caption{Qualitative results of generative identity unlearning on AFHQv2-Cat dataset.}
    \label{fig:cat}
\end{figure}
\begin{table}[!t]
\centering
\resizebox{0.7\columnwidth}{!}{%
\begin{tabular}{c|cc}
\toprule[1.1pt]
 & FID\textsubscript{pre} ($\downarrow$) & $\Delta$FID\textsubscript{real} ($\downarrow$) \\ \hline
AFHQv2-Cat & 5.93 \scriptsize{$\pm$ 1.03} & 3.44 \scriptsize{$\pm$ 1.68} \\ \bottomrule[1.1pt]
\end{tabular}%
}
\caption{Quantitative results of generative identity unlearning on AFHQv2-Cat dataset. The existing ID metrics are designed for the human face, and there are no adequate metrics for cat. For this reason, we only represent about FID\textsubscript{pre} and $\Delta$FID\textsubscript{real} in this experiments.}
\vspace{-0.5cm}
\label{tab:cat}
\end{table}
\subsection{Generative Identity Unlearning in AFHQv2}
In this section, we validated \frameworkname{} in a different dataset - AFHQv2-Cat \cite{choi2020starganv2}.
We used the generator architecture \cite{chan2022efficient} and the GAN inversion network \cite{yuan2023make} pre-trained on AFHQv2-Cat.
The pre-trained weights are publicly available at their official implementations.
Since the identity loss \cite{deng2019arcface} used in our main experiment were designed to capture the dissimilarities between identities in human faces, we only adopted to use the reconstruction loss and the perceptual loss \cite{zhang2018unreasonable} in this experiment.
The qualitative results are shown on Figure \ref{fig:cat}.
We could show the effectiveness of \frameworkname{} in a different domain - faces of cats.
In Table \ref{tab:cat}, we additionally show that GUIDE can preserve performance of pre-trained model on AFHQv2-Cat.

\begin{figure*}[!h]
    \centering
    \includegraphics[width=0.8\textwidth]{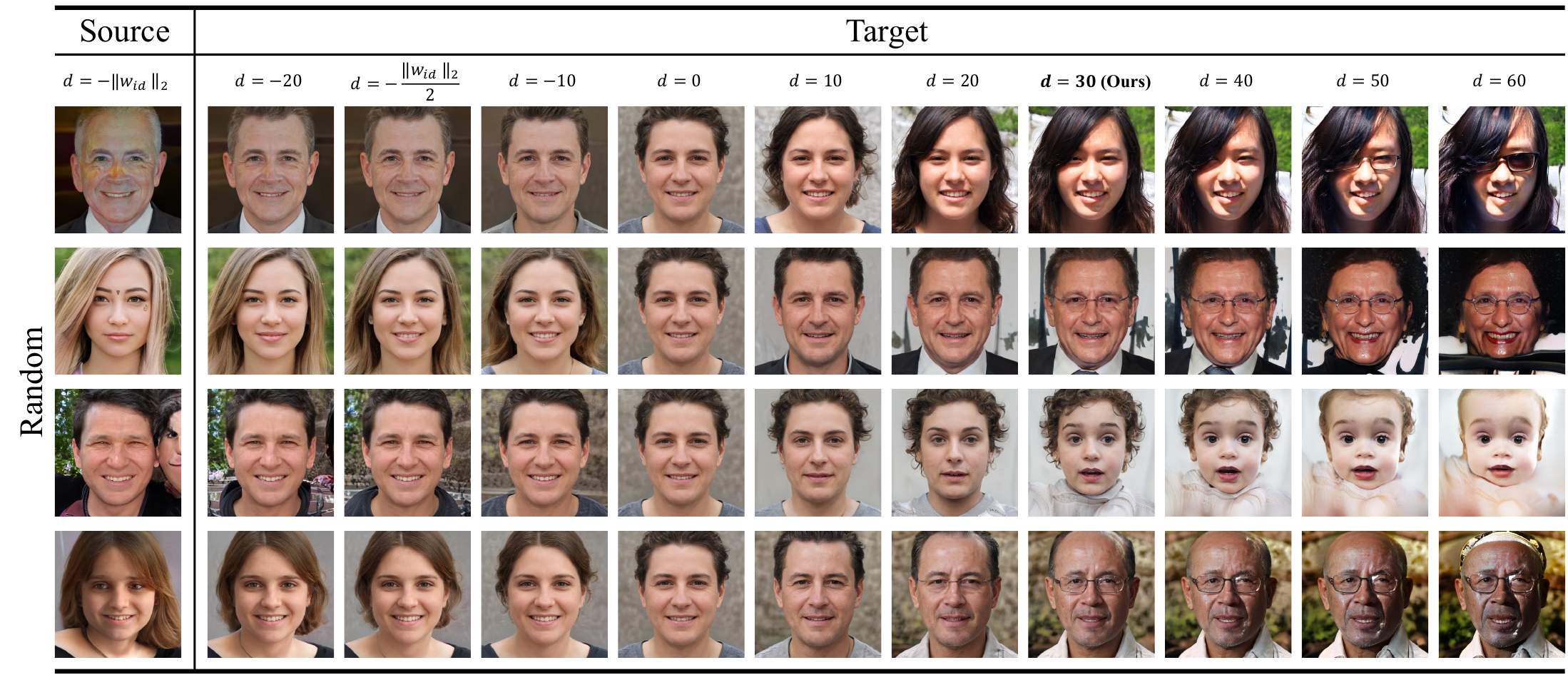}
    \caption{Illustration of target images from source images with different $d$ in \textit{Random} scenario.}
    \label{fig:multi_source_distance}
\end{figure*}
\begin{figure*}[!h]
    \centering
    \includegraphics[width=0.8\textwidth]{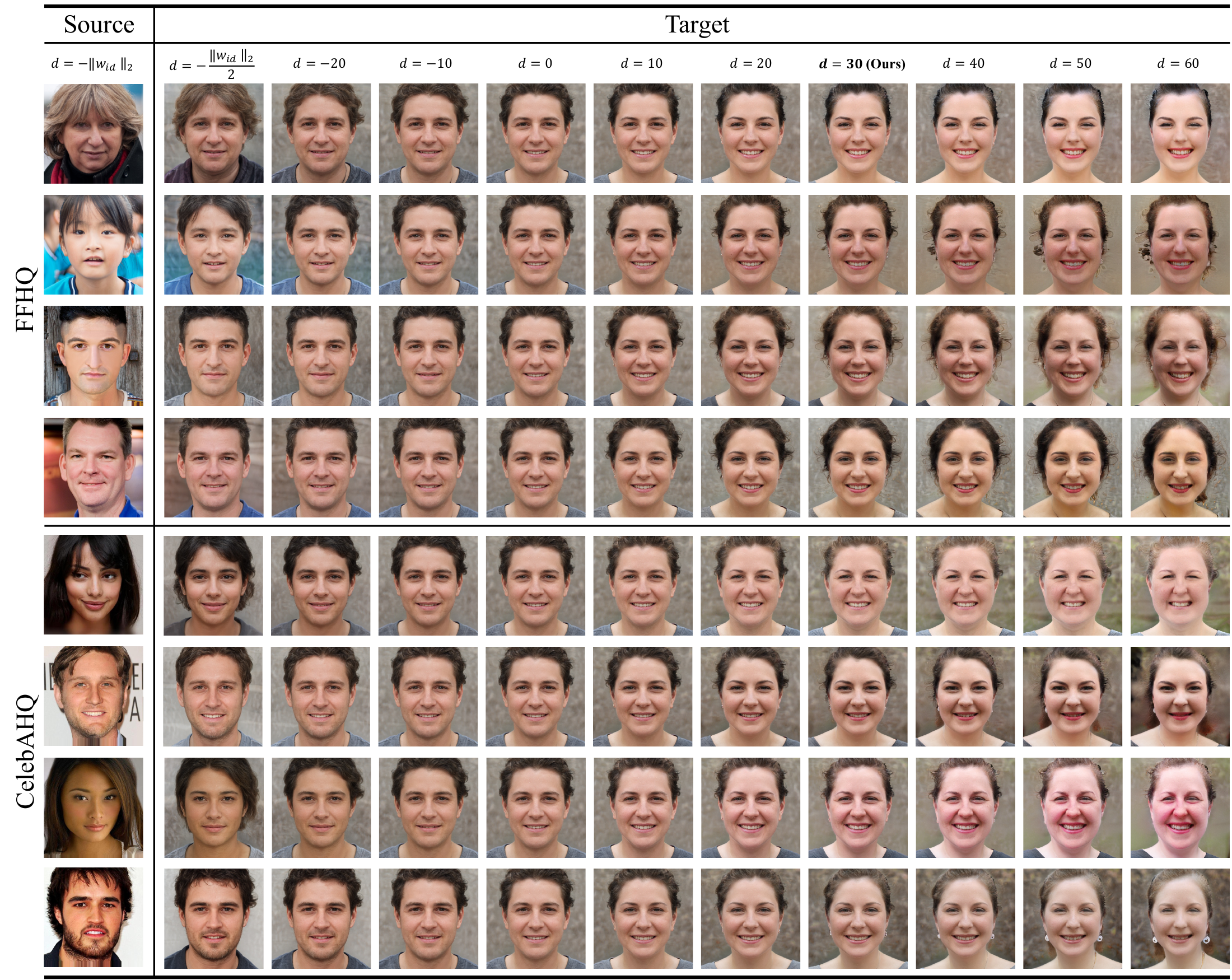}
    \caption{Illustration of target images from source images with different $d$ in \textit{In-domain} (FFHQ) and \textit{Out-of-domain} (CelebAHQ) scenario.}
    \label{fig:multi_source_distance_2}
\end{figure*}
\begin{figure*}[!h]
    \centering
    \includegraphics[width=0.8\textwidth]{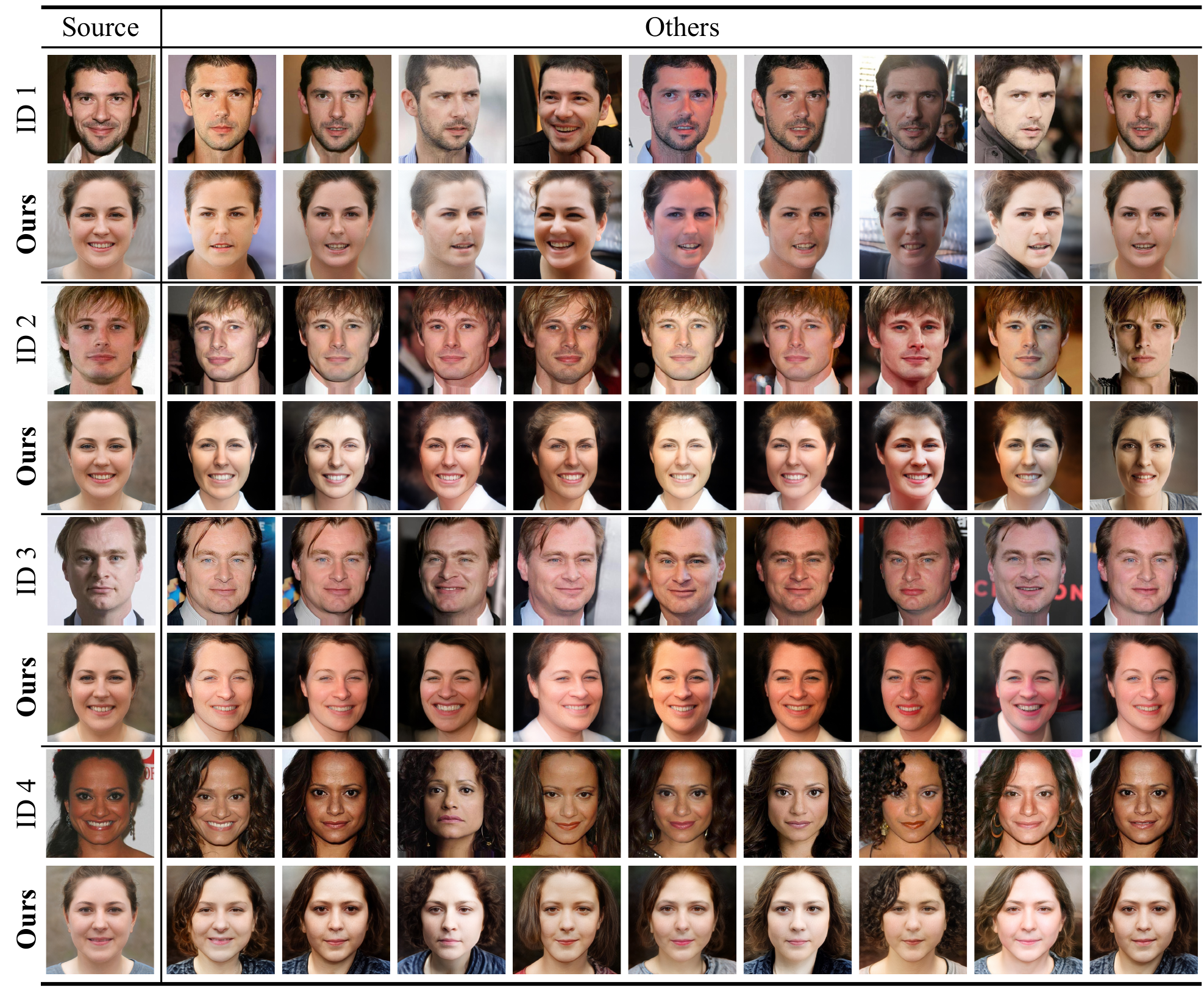}
    \caption{Additional qualitative results with CelebAHQ dataset.}
    \label{fig:additional_qual}
\end{figure*}
\begin{figure*}[!h]
    \centering
    \includegraphics[width=0.65\textwidth]{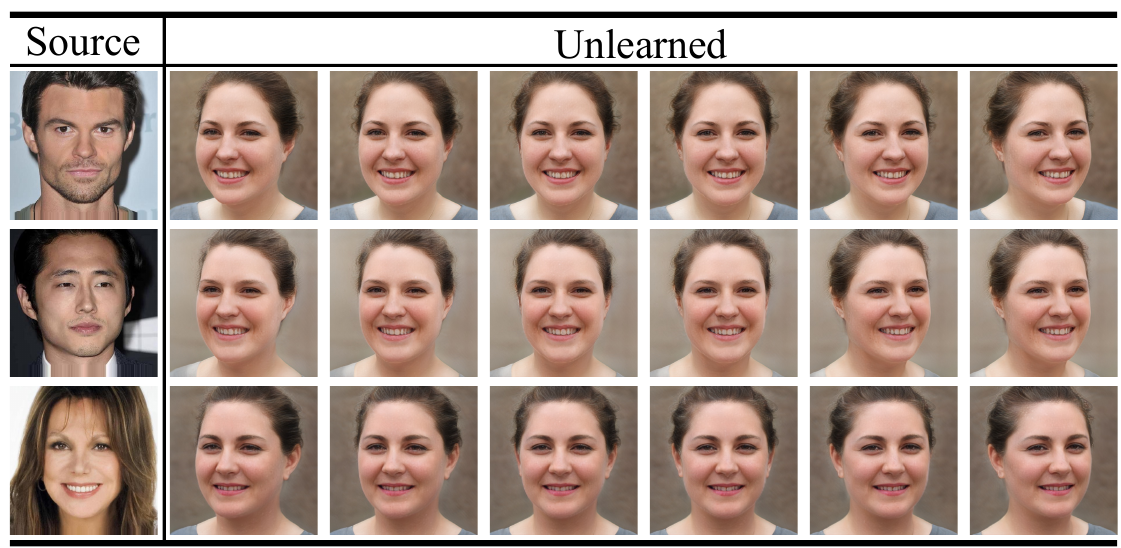}
    \caption{Generated images given the source image with different camera poses.}
    \vspace{-0.3cm}
    \label{fig:multi_view}
\end{figure*}
\subsection{Target Images from Different \texorpdfstring{$\bm{d}$}{TEXT}}
In addition to Section 4.3 in our main paper, titled ``Effect of $d$ in Determination of $w_t$", this section presents further experiments involving a variety of target images.
We visualized target images derived from a given source image at multiple $d$ values.
In Figure \ref{fig:multi_source_distance}, we utilized various $d$ values to sample the corresponding target image in the \textit{Random} scenario, while Figure \ref{fig:multi_source_distance_2} is for the \textit{In-domain} and \textit{Out-of-domain} scenarios.
Our results illustrate that adjusting $d$ allows us to obtain diverse target images.
However, as mentioned in our main paper, target images derived from interpolated latent code, where $d$ is less than 0, exhibit similarity to the given source image.
Conversely, target images with $d\geq50$ tend to be corrupted.
Therefore, our choice of $d=30$ appears to strike a visually balanced representation for the target image.

\subsection{Additional Qualitative Results}
In this section, we presented additional qualitative results.
In contrast to our main paper, we utilized 10 images per identity, and the results are illustrated in Figure \ref{fig:additional_qual}.
The findings emphasize once again that \frameworkname{} successfully erases the identity not only in the given source image but also in other images with the same identity.

\subsection{Multi-View Synthesized Images}
\vspace{-0.15cm}
In this section, we visualized the unlearned images from continuous camera poses.
We conducted this experiment within \textit{Out-of-domain} scenario.
As shown in Figure \ref{fig:multi_view}, our unlearning process successfully erased the source identity across multiple poses.

\begin{table}[!t]
\centering
\resizebox{\columnwidth}{!}{%
\begin{tabular}{l|ccc|ccc}
\toprule[1.1pt]
\multirow{2}{*}{Method} &
  \multicolumn{3}{c|}{In-Domain (FFHQ)} &
  \multicolumn{3}{c}{Out-of-Domain (CelebAHQ)} \\ \cline{2-4}\cline{5-7}
 &
  ID  &
  FID\textsubscript{pre} &
  $\Delta$FID\textsubscript{real} &
  ID&
  FID\textsubscript{pre} &
  $\Delta$FID\textsubscript{real}\\ \hline
  Baseline & 0.14  & 8.60  & 5.97
           & 0.05   & 6.75  & 4.32
           \\
\textbf{Ours} & \textbf{0.06} & \textbf{6.14} & \textbf{4.35} 
           & \textbf{0.01} & \textbf{6.07} & \textbf{4.25}     \\ \bottomrule[1.1pt]
\end{tabular}%
}
\caption{Quantitative results of \frameworkname{}-SG2 (Ours) and the baseline in the \taskname{} task, tested in a single-image setting using
one image per identity. Due to space limit, we presented the corresponding standard deviations in the Supplementary materials.}
\label{tab:single_image_sg2}
\end{table}
\begin{figure}[!t]
    \centering
    \includegraphics[width=\columnwidth]{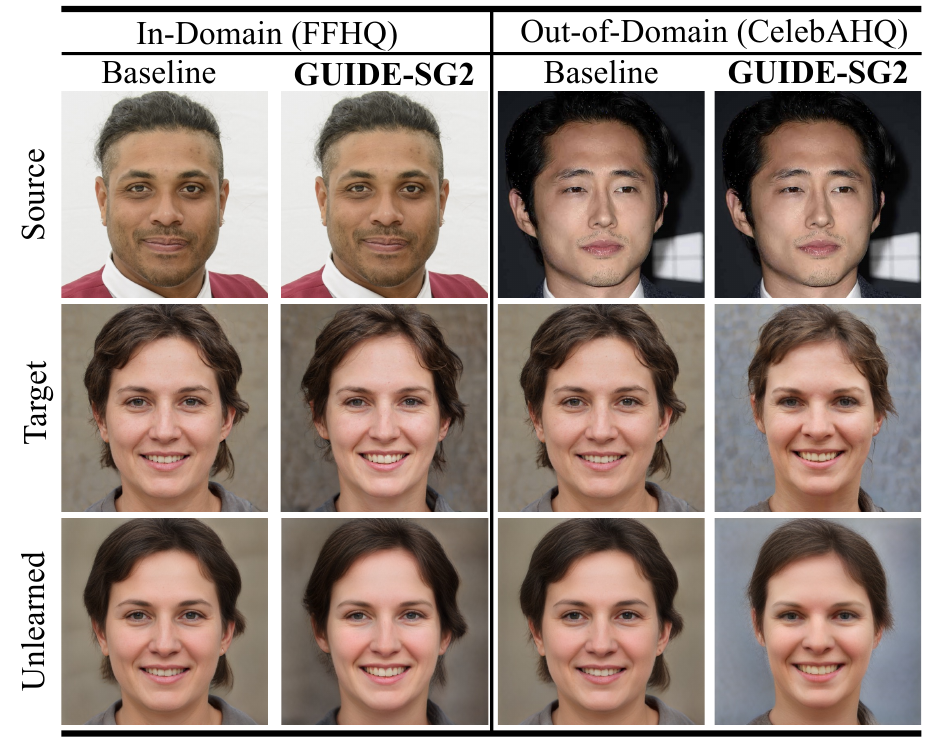}
    \caption{Qualitative results of \frameworkname{}-SG2 and the baseline. For the given source image each (the first
    row), \frameworkname{}-SG2 and the baseline tried to erase the identity in the pre-trained generator. The result are shown on the second row. Images in the third row are the target image in our unlearning process.}
    \label{fig:single_image_sg2}
\end{figure}
\vspace{0.2cm}
\subsection{Generative Unlearning in StyleGAN2}
In addition to our primary experiments employing a 3D generative adversarial network as the generator architecture, we observed the effectiveness of our framework in unlearning identity in 2D generative adversarial networks.
In this section, we utilized the widely-used StyleGAN2 \cite{karras2020analyzing} as the backbone architecture and pSp \cite{richardson2021encoding} as a GAN inversion network for latent code extraction from images.
Both the backbone and the GAN inversion network were pre-trained on FFHQ \cite{karras2020analyzing}.
We refer to our framework built on top of StyleGAN2 as \frameworkname{}-SG2.

In \frameworkname{}-SG2, we employed images from the StyleGAN2 generator for calculating loss, instead of tri-plane feature maps.
We present results of \frameworkname{}-SG2 qualitatively in Figure \ref{fig:single_image_sg2} and quantitatively in Table \ref{tab:single_image_sg2}.
Both results demonstrated \frameworkname{}-SG2  successfully erased the given identity in a 2D GAN architecture with minimal impact on the performance of the pre-trained generator.
\section{Additional Ablation Study}
\subsection{Number of Latent Codes in Loss Functions}
\begin{table}[!t]
\centering
\resizebox{\columnwidth}{!}{%
\begin{tabular}{c|cccc}
\toprule[1.1pt]
$N_a$ & ID ($\downarrow$) & ID\textsubscript{others} ($\downarrow$) & FID\textsubscript{pre} ($\downarrow$) & $\Delta$FID\textsubscript{real} ($\downarrow$) \\ \hline
1 & 0.046 \scriptsize{$\pm$ 0.054} & 0.191 \scriptsize{$\pm$ 0.077} & 8.001 \scriptsize{$\pm$ 1.992} & 3.515 \scriptsize{$\pm$ 1.153}\\
\textbf{2} & \textbf{0.030 \scriptsize{$\pm$ 0.051}} & \textbf{0.174 \scriptsize{$\pm$ 0.081}} & 7.882 \scriptsize{$\pm$ 1.958} & 3.442 \scriptsize{$\pm$ 1.104} \\
4 & 0.034 \scriptsize{$\pm$ 0.055} & 0.184 \scriptsize{$\pm$ 0.077} & \textbf{7.668 \scriptsize{$\pm$ 1.807}} & \textbf{3.340 \scriptsize{$\pm$ 1.028}} \\ \bottomrule[1.1pt]
\end{tabular}%
}
\caption{Ablation study to figure out the optimal $N_a$. 
To find optimal, we performed the analysis with different $N_a$. As can be seen, when $N_a$ is 2, GUIDE erase the identity most effectively, and the performance of the pre-trained model can be preserved. We used CelebAHQ dataset in this experiment.}
\label{tab:ablation_n_a}
\end{table}
\begin{table}[!t]
\centering
\resizebox{\columnwidth}{!}{%
\begin{tabular}{c|cccc}
\toprule[1.1pt]
$N_g$ & ID ($\downarrow$) & ID\textsubscript{others} ($\downarrow$) & FID\textsubscript{pre} ($\downarrow$) & $\Delta$FID\textsubscript{real} ($\downarrow$) \\ \hline
1 & 0.065 \scriptsize{$\pm$ 0.067} & 0.180 \scriptsize{$\pm$ 0.080} & 7.875 \scriptsize{$\pm$ 2.017} & 3.491 \scriptsize{$\pm$ 1.118} \\
\textbf{2} & \textbf{0.030 \scriptsize{$\pm$ 0.051}} & \textbf{0.174 \scriptsize{$\pm$ 0.081}} & 7.882 \scriptsize{$\pm$ 1.958} & 3.442 \scriptsize{$\pm$ 1.104} \\
4 & 0.031 \scriptsize{$\pm$ 0.055} & 0.181 \scriptsize{$\pm$ 0.075} & \textbf{7.705 \scriptsize{$\pm$ 1.868}} & \textbf{3.359 \scriptsize{$\pm$ 1.079}}\\ \bottomrule[1.1pt]
\end{tabular}%
}
\caption{Ablation study to figure out the optimal $N_g$. 
To find optimal, we performed the analysis with different $N_g$. As can be seen, when $N_g$ increase, GUIDE can preserve the performance of pre-trained model more effectively. When $N_g=2$, GUIDE have achieved a balanced performance in our metric. We used CelebAHQ dataset in this experiment.}
\label{tab:ablation_n_g}
\end{table}
In the computation of $\mathcal{L}_{adj}$ and $\mathcal{L}_{global}$, as outlined in Section 3.3 of our main paper, we incorporated $N_a$ and $N_g$ latent codes, respectively.
In this section, we investigate the influence of varying $N_a$ and $N_g$.
Due to an out-of-memory issue in VRAM, these experiments were conducted on an NVIDIA A6000 GPU.
Table \ref{tab:ablation_n_a} presents the results of varying $N_a$ in $\mathcal{L}_{adj}$ while keeping $N_g$ fixed at 2.
Our findings indicate that using $N_a=2$ yields the best performance in erasing the given identity among different values of $N_a$, while maintaining comparable performance in preserving generation quality.
Conversely, in Table \ref{tab:ablation_n_g}, we varied $N_g$ in $\mathcal{L}_{global}$ while keeping $N_a$ fixed at 2.
Results show that using $N_g=2$ achieves a balanced performance between erasing the given identity and preserving generation performance.
Importantly, all cases experimented upon outperformed the baseline in \taskname{} task.

\begin{table}[!t]
\centering
\resizebox{\columnwidth}{!}{%
\begin{tabular}{c|cccc}
\toprule[1.1pt]
$\lambda_{L2}$ & ID ($\downarrow$) & ID\textsubscript{others} ($\downarrow$) & FID\textsubscript{pre} ($\downarrow$) & $\Delta$FID\textsubscript{real} ($\downarrow$) \\ \hline
$10^{-3}$ & 0.089 \scriptsize{$\pm$ 0.060} & 0.269 \scriptsize{$\pm$ 0.086} &  \textbf{4.815 \scriptsize{$\pm$ 1.000}} & \textbf{1.454 \scriptsize{$\pm$ 0.294}} \\
$\mathbf{10^{-2}}$ & \textbf{0.030 \scriptsize{$\pm$ 0.051}} & 0.174 \scriptsize{$\pm$ 0.081} & 7.882 \scriptsize{$\pm$ 1.958} & 3.442 \scriptsize{$\pm$ 1.104} \\
$10^{-1}$ & 0.032 \scriptsize{$\pm$ 0.053} & \textbf{0.159 \scriptsize{$\pm$ 0.073}} & 13.308 \scriptsize{$\pm$ 2.989} & 7.783 \scriptsize{$\pm$ 1.808}\\
1 & 0.036 \scriptsize{$\pm$ 0.052} & 0.161 \scriptsize{$\pm$ 0.073} & 15.034 \scriptsize{$\pm$ 3.079} & 9.198 \scriptsize{$\pm$ 1.858} \\ \bottomrule[1.1pt]
\end{tabular}%
}
\caption{Ablation study to figure our the optimal $\lambda_{L2}$. We compared the performance among different $\lambda_{L2}$ in CelebAHQ dataset.}
\label{tab:ablation_lambda_l2}
\end{table}
\begin{table}[!t]
\centering
\resizebox{\columnwidth}{!}{%
\begin{tabular}{c|cccc}
\toprule[1.1pt]
$\lambda_{id}$ & ID ($\downarrow$) & ID\textsubscript{others} ($\downarrow$) & FID\textsubscript{pre} ($\downarrow$) & $\Delta$FID\textsubscript{real} ($\downarrow$) \\ \hline
$10^{-2}$ & 0.033 \scriptsize{$\pm$ 0.053} & 0.177 \scriptsize{$\pm$ 0.080} & \textbf{7.879 \scriptsize{$\pm$ 1.943}} & 3.460 \scriptsize{$\pm$ 1.093} \\
$\mathbf{10^{-1}}$ & \textbf{0.030 \scriptsize{$\pm$ 0.051}} & \textbf{0.174 \scriptsize{$\pm$ 0.081}} & 7.882 \scriptsize{$\pm$ 1.958} & \textbf{3.442 \scriptsize{$\pm$ 1.104}} \\
1 & 0.105 \scriptsize{$\pm$ 0.053} & 0.244 \scriptsize{$\pm$ 0.081} & 7.920 \scriptsize{$\pm$ 1.781} & 3.534 \scriptsize{$\pm$ 0.953} \\ \bottomrule[1.1pt]
\end{tabular}%
}
\caption{Ablation study to figure our the optimal $\lambda_{id}$. We compared the performance among different $\lambda_{id}$ in CelebAHQ dataset.}
\label{tab:ablation_lambda_id}
\end{table}
\subsection{Scaling Factors of Loss Functions}
In $\mathcal{L}_{local}$ and $\mathcal{L}_{adj}$, as proposed in Section 3.3 of our main paper, we set the scaling factors as $\lambda_{L2}=10^{-2}$ and $\lambda_{id}=10^{-1}$.
In this section, we conducted ablation studies to determine the effective scaling factors.

In Table \ref{tab:ablation_lambda_l2}, we varied $\lambda_{L2}$ while fixing the other scaling factors as default.
For small $\lambda_{L2}$, the generator architecture could not successfully erase the given identity.
On the other hand, for larger $\lambda_{L2}$, the generator architecture lost generation performance significantly.
Based on these observations, we decided to use $\lambda_{L2}=10^{-2}$ for balanced performance.
In Table \ref{tab:ablation_lambda_id}, we varied $\lambda_{id}$ while fixing the other scaling factors as default.
Our findings indicate that using $\lambda_{id}=10^{-1}$ is the most effective.

\section{Additional Implementation Details}
Besides the Section 4.1 in our main paper, in this section, we additionally describe the implementation details that are omitted in the main paper due to space limit.
We ran \frameworkname{} and the baseline using a single NVIDIA A5000 GPU.
Erasing the given source identity using \frameworkname{} takes about 20 minutes.
We utilized only a single image to represent a certain identity; \frameworkname{} underwent 1,000 iterations throughout our experiments.
{
    \small
    \bibliographystyle{ieeenat_fullname}
    \bibliography{main}
}


\end{document}